\documentclass[USenglish,oneside,twocolumn]{article}

\usepackage[utf8]{inputenc}
\usepackage[big]{dgruyter_NEW}

\usepackage{color}
\usepackage{amssymb}
\usepackage{subcaption}

\newcommand\xxx[2][Name]{\textbf{[\textcolor{red}{#2} -- {#1}]}}
\renewcommand\xxx[2][Name]{}
 
\newcommand\mr[1]{\textcolor{blue}{#1}}
\renewcommand\mr[1]{#1}
 
\newcommand\heading[1]{\vspace*{4pt}\noindent \textbf{#1.}}  

\begin{document}

  \author*[1]{Ivan Evtimov}

  \author[2]{Pascal Sturmfels}

  \author[3]{Tadayoshi Kohno}


  \affil[1]{Paul G. Allen School of Computer Science \& Engineering, University of Washington, E-mail: ie5@cs.washington.edu}

  \affil[2]{Paul G. Allen School of Computer Science \& Engineering, University of Washington}

  \affil[3]{Paul G. Allen School of Computer Science \& Engineering, University of Washington}


  \title{\huge FoggySight: A Scheme for Facial Lookup Privacy}

  \runningtitle{FoggySight}


  \begin{abstract}
{
Advances in deep learning algorithms have enabled better-than-human performance on face recognition tasks.
In parallel, private companies have been scraping social media and other public websites that tie photos to identities and have built up large databases of labeled face images.
Searches in these databases are now being offered as a service to law enforcement and others and carry a multitude of privacy risks for social media users. 
In this work, we tackle the problem of providing privacy from such face recognition systems.
We propose and evaluate FoggySight, a solution that applies lessons learned from the adversarial examples literature to modify facial photos in a privacy-preserving manner before they are uploaded to social media. 
FoggySight's core feature is a community protection strategy where users acting as protectors of privacy for others upload decoy photos generated by adversarial machine learning algorithms.
We explore different settings for this scheme and find that it does enable protection of facial privacy -- including against a facial recognition service with unknown internals.
}
\end{abstract}



\maketitle
\section{Introduction}
Owing to advances in deep learning (DL), face verification and recognition have made tremendous strides in the past decade.
A report from 2018 by the National Institute of Standards and Technologies (NIST) concluded that models can identify the correct identity of an individual in a photo from a set of 12 million identities with error rates less than 0.2\%~\cite{grother2018ongoing}, far surpassing human performance~\cite{kemelmacher2016megaface}.
As a result, face recognition is now being successfully applied for device authentication, identity verification, and payment authorization. 

Unfortunately, this progress has also enabled unprecedented invasions of individual privacy.
Media reporting has revealed that private companies are applying facial recognition technology to link photos of individuals in public places to their social media identities~\cite{hill_valentino-devries_dance_krolik_2020,heilweil2020vox}.
This is achieved by building up a database of facial photos associated with profiles on web sites such as Facebook, Twitter, and even Venmo~\cite{heilweil2020vox}.
A photo taken from anyone anywhere can then be processed by a DL face recognition model to match it up with the photos in the database.
At least one service is currently being pioneered as a face search database for law enforcement agencies and raises a host of civil liberties questions around involuntary inclusion in criminal databases and reasonable search~\cite{eff2018facereco}.
Other similar services also exist~\cite{chivers2019facewatch}.
It is also easy to imagine more nefarious applications of this easily accessible technology. 
Stalkers, who have only seen potential victims online, could apply this technology to identify individuals in public web cam video streams; 
see~\cite{steadman2020stalkers} for a motivating example.
Criminal or other illegal organizations could also use this technology to identify people in news media photos and then target those people for physical harm or retaliation; see~\cite{owen2020doxxing} for a motivating example. 
These examples illustrate that individuals uploading pictures to social media websites are exposing themselves to the risk of future identification in new photos via DL-enabled facial searches.

Any solution to protect individual privacy must acknowledge these realities: that \mr{facial search databases already contain previously} publicly available \mr{tagged} photos of many (possibly millions of) individuals, that individuals cannot predict when they are at risk of being targeted by a face recognition system, and those photos being used for face recognition may come from sources external to the social media platforms. 
In this work, we propose a new framework for protecting against face recognition that takes these issues into account. 
We propose using adversarial examples --- small perturbations to images that fool DL models but are imperceptible to humans --- \mr{to poison the lookup database of facial search services}. 
This involves coordination of adversarial modifications among many users: a large number of adversarial photos uploaded by many different individuals may protect privacy \mr{by ``crowding out'' previously scraped ``clean'' photos of individuals in response to queries} \emph{without} those individuals needing to obscure their identity when in public.

\mr{
    Our contributions are as follows:
    \begin{itemize}
        \item We propose FoggySight: a collaborative facial privacy approach meant to poison the database used for facial search. We study the conditions needed for FoggySight to be successful and find that individuals can meaningfully increase their privacy when other ``protectors'' feed adversarially modified photos (``decoys'') in the facial database.
        \item We compare and evaluate different approaches for generating adversarial examples/decoys in the metric learning space defined by face recognition neural networks and find the most effective approach to be to target the mean of an individual's available facial vectors.
        In that scenario, protected individuals only need protectors to provide decoys numbering 2-4 times the number of unmodified photos of the protected, when protectors have access to the facial search model.
        \item When protectors do not have access to the facial search model, they need to increase both the magnitude of modifications in the decoys and the number they provide relative to the clean photos of the protected.
        But they can still meaningfully increase the privacy of the protected: under the right parameters, we show that our scheme can decrease the identification rate on the Azure Face Service to under 10\%.
    \end{itemize}
}

\mr{
    We emphasize that our work is meant to advance the exploration of facial privacy protections with community methods and adversarial examples but that more thorough evaluations are needed for final deployment of this scheme.
    We discuss limitations in detail in Section~\ref{sec:discussion}.
}

\section{Background}
\label{sec:background}
\subsection{Face Recognition and Terminology}
\label{sec:facerecognition}
Automated face recognition has had a long history in the computer vision community \citep{bruce1986understanding}.
Some of the earliest approaches to face recognition made use of basis decompositions \citep{turk1991face, he2005face}, local binary patterns \citep{ahonen2006face} and SIFT features \citep{bicego2006use}. 
More recent approaches have made use of deep neural networks to automatically classify faces into known identities \cite{taigman2014deepface, sun2014deep}. 
These approaches are limited to only being able to classify faces from a known, preset list (e.g., the faces the model was trained on).
To overcome this, state of the art approaches have cast face recognition as a metric learning problem. 
In this view, the goal is to learn an embedding space in which two faces of the same person are close and two faces of different people are far away. 
There exist many proposed loss functions to learn such an embedding space, including paired \cite{sun2015deeply, hu2014discriminative} and triplet losses \cite{schroff2015facenet, parkhi2015deep, hermans2017defense} --- which directly optimize distance between pairs of faces --- and clustering or max-margin style losses \cite{wen2016discriminative, liu2017sphereface, wang2018cosface, deng2019arcface}, which aim to classify faces with an additive or multiplicative margin.

\begin{figure}[tb]
    \centering
    \includegraphics[width=0.9\columnwidth]{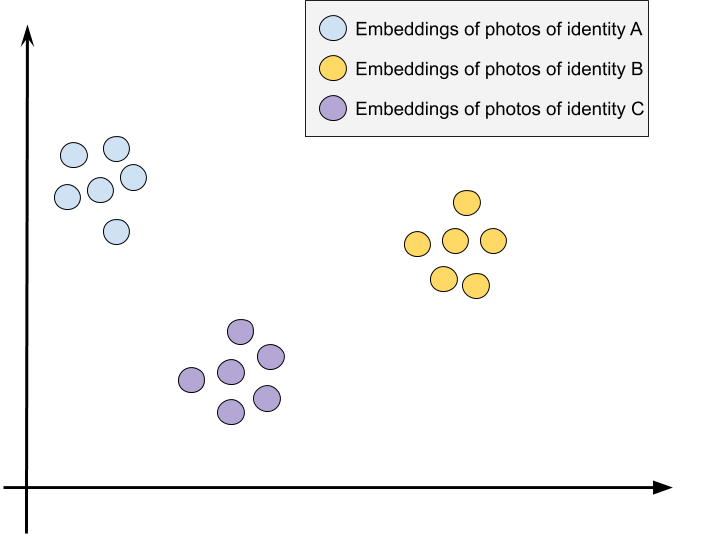}
    \caption{ 
    Simplified visual representation of the metric space learned by state-of-the-art face recognition neural networks.
    When a tightly cropped facial image is processed by a neural network, it produces an embedding vector (here, represented by a dot in $\mathbb{R}^2$). 
    Pairs of vectors belonging to different identities are far away from each other while those belonging to the same identity are close together.
    In practice, neural networks produce vectors in $\mathbb{R}^{128}$ or $\mathbb{R}^{512}$ and metrics such as Euclidean distance and cosine similarity define ``closeness.'' 
    }
    \label{fig:neural_face_embeddings}
\end{figure}

This more modern paradigm of metric learning differs from traditional classification in that the neural network models don't produce direct identity predictions. 
Rather, they produce embedding vectors of each input image such that images belonging to a given identity are clustered in the embedding space (see Fig.~\ref{fig:neural_face_embeddings}). 
This allows rapid face verification and lookup for identities that are not necessarily included in the network's training set via $k$-nearest neighbors. 

A modern pipeline for face recognition using such a neural network might look as follows. First, the face recognition company either downloads a pre-trained, publicly available neural network designed for face recognition, or trains one themselves on an existing dataset where the identities are labeled. Then, they scrape the internet for publicly available photos from social media. They release an application combining their dataset and network. A user of the app takes a photo of a stranger in public. That photo is uploaded to the face recognition company's server, where it is cross-referenced against the photos collected from social media websites. The most similar faces according to the neural network are returned to the user of the app, along with the associated social media profiles. 
This is the approach used by the companies described in~\cite{heilweil2020vox}.

To aid with further discussion of this pipeline, we introduce several terms and notations. We denote the face recognition model as $f:\mathbb{R}^{w \times h \times c} \mapsto \mathbb{R}^d$ for some latent embedding space of dimension $d$ and images of size $w \times h \times c$. In this embedding space, similarity between two faces is computed using normalized distance in the embedding space. That is, for two images $x_1$, $x_2$, the distance function $D$ between them is evaluated as:

\[D(x_1, x_2) = \left|\left|\frac{f(x_1)}{||f(x_1)||_2} - \frac{f(x_2)}{||f(x_2)||_2}\right|\right|_2^2\]

In addition, we define the following terms:
\begin{itemize}
    \item \textbf{Lookup Set}: The set of photos that a face recognition company scrapes from social media. 
    These photos, along with their associated profiles or links, are those that are cross-referenced against when identifying an individual in a new photo. 
    Each photo in the lookup set represents an embedding vector in the neural network's output space -- the nearest lookup set photos to the query photo are returned when performing a search. We denote the lookup set by $L$. 
    \item \textbf{Query Photo}: The photo that the user of face recognition technology (the adversary in our model, see Section~\ref{sec:goals_and_assumptions}) wants to match to an identity. 
    This photo may be a photo  of, for example, a stranger in a public place. 
    This photo is processed by the neural network and a vector is produced in the embedding space that can be compared against the vectors of the lookup set photos.
    The closest neighbors of the lookup set photos in embedding space are returned as candidate matches.
    \item \textbf{Top $k$ Recall Set} The set of $k$ closest neighbors in the lookup set to the embedding vector corresponding to the query photo. $k$ is a parameter that can be adapted for broader or narrower searches. For some query photo $q$, lookup set $L$ and distance metric $D$, we use the following notation to denote this set:
    \[N(q, k) := \textrm{arg top-k}_{x \in L} (D(q, x)) \]
    Some facial recognition services -- such as the Microsoft Azure Face Service that we study -- only expose $N(q, 1)$ in their API responses.
\end{itemize}

\subsection{Adversarial Examples and Face Recognition}

There exist many works demonstrating the vulnerability of deep learning face recognition systems to \textit{adversarial examples}.
Adversarial examples were first discovered by~\cite{szegedy2013intriguing}, who noticed that certain small-magnitude modifications to images shift the output of neural networks in unexpected ways. 
Since that first publication, a large body of literature has designed strong algorithms for generating such adversarial images that circumvent even state-of-the-art defenses and remain indistinguishable from benign images for humans~\citep{carlini2017towards, carlini2017adversarial, athalye2018obfuscated, tramer2020adaptive}.
Other works have established that adversarial examples can transfer between models~\cite{liu2016delving} and that they can be generated without access to the internals of the model~\cite{papernot2017practical}.
In short, adversarial examples are fundamental vulnerabilities in DL models that have not been remedied reliably to this day.
\mr{They allow their creators to control the output of neural network models while preserving visual similarity to non-adversarial images.}

\mr{One set of works seeking to fool facial recognition models has focused on creating physical adversarial examples in the form of objects -- such as glasses frames~\cite{sharif2016accessorize, sharif2019general} and hats~\cite{komkov2019advhat} -- that change the output of a model processing images of a person wearing them.}
Others showed that generating adversarial examples is also possible without possession of the weights and biases of the neural network but only with query-based oracle-like access to the model~\cite{dong2019efficient}.
\mr{
    Of particular interest is work by \citet{gao2020face} that develops transferable adversarial examples by optimizing in the metric space of facial recognition networks and studies how much distortion individuals are willing to accept in their photos.
    In addition, \citet{rajabi2021practicality} and \citet{oh2017adversarial} develop new approaches for generating adversarial examples againt facial recognition that do not rely on the ``standard'' methods from~\cite{carlini2017towards,madry2017towards} and show that they are robust even in the face of countermeasures.
}

These works certainly indicate that adversarial examples and adversarial objects are particularly attractive mechanisms for protecting privacy from facial recognition.
However, if individuals are to act as ``attackers'' of the neural network 
under the assumptions of the literature so far, they should be able to modify \mr{the query photo used to perform the search}, or change their own appearance permanently. 
Neither of these is possible in a real-world scenario.
Individuals can hardly control the photos others take of them.
Anybody can snap a picture of anybody in a public space and CCTV and well-meaning web cams are pervasive.
Furthermore, wearing adversarial accessories -- such as hats and glasses -- is not always practical or fashionable and restricts the individual's freedom to control their own appearance. 
This is why we explore a scheme that does not assume control of the photo used to de-anonymize the individual.

\mr{
    A concurrent conference submission by~\citet{shan2020fawkes} explores a similar solution to ours. 
    This proposal, named Fawkes, also uses adversarial examples -- named ``cloaks'' in that work -- to disrupt the performance of facial classification.
    Cloaks have the same purpose as decoys in our work and are like adversarial examples from the adversarial machine learning literature.
    The authors also discuss a ``Sybil attack'' which corresponds to our communal defense strategy in which protector users upload cloaks/decoys/adversarial examples modified so that facial recognition models output a vector or classification corresponding to another user.
    Our work adds additional perspective by exploring what vector targets are best to use by the protectors and by applying an alternative transferable adversarial examples generation mechanism.
    In Section~\ref{sec:selecting_targets}, we propose a number of possible mechanisms to select vectors in the metric learning space to use as targets for protectors and discuss their tradeoffs; in Section~\ref{sec:evaluation}, we evaluate and compare those different approaches quantiatively.
    Furthermore, transferable cloak generation in Fawkes requires the protectors to use a robust neural network model trained on adversarial examples.
    In Section~\ref{sec:advexgen}, we discuss an alternative method that does not require retraining and uses ``out-of-the-box'' models available online; we evaluate this method and find it to be successful in Section~\ref{sec:transferability}.
    A tradeoff of our method relative to Fawkes is that it requires larger perturbations to achieve privacy protections. Together, \citet{shan2020fawkes} and our paper provide a robust foundation for protecting face recognition under our shared threat model.
}

\mr{
  More broadly, our line of inquiry also fits in with studies on applying adversarial machine learning for beneficial goals, such as~\cite{delobelle2020ethical, volkel2020trick, albert2020politics, howe2017engineering}. 
  In most adversarial machine learning research, the party performing adversarial modifications is often referred to as the adversary. 
  However, looking toward Section~\ref{sec:goals_and_assumptions}, we note  that in our work -- as in these other works -- the party performing these modifications is not the adversary, but the party seeking defense against an adversary. 
  Because FoggySight's participants perform these ``attacks'' against the adversary's capabilities, we may sometimes use the word ``attack'' to refer to the actions of the privacy protectors.
}

\subsection{Other Attacks on Face Recognition Models}
There exists a limited amount of work that attempts to fool face recognition systems by modifying photos at \textit{training time} rather than at test time. 
\citet{chen2017targeted} introduce a set of data poisoning attacks that modify a small number of the training photos in a face dataset. 
They show that a model trained on the poisoned dataset learns a back-door key: a pattern that, when presented to the model, gets the model to categorize that pattern as belonging to a particular face for impersonation purposes. 
They further demonstrate that they can instantiate this back-door key in the physical world by making the learned pattern a specific pair of glasses. 
Not specific to face recognition, there exists a body of work on attacking neural network systems using data poisoning attacks \citep{liu2017trojaning, gu2017badnets, shafahi2018poison, munoz2017towards}.\\

There exists a subset of work on designing face recognition systems to be private \citep{erkin2009privacy, sadeghi2009efficient, masi2018deep, xiang2016privacy}.
\mr{Those, in turn, are similar to work aiming to preserve the privacy of training set members and individual features of training set examples in machine learning more broadly~\citep{jia2018attriguard,jia2019memguard}.} 
These works aim to design machine learning systems that don't expose the model or database or (features of the) training set to the user and don't expose the user to the model or server running the model. 
We view these works as tangential to ours: they still aim to design systems that are fundamentally able to identify individuals. 
Our main goal is to thwart such systems, with the assumption that those employing face recognition technology are not interested in our privacy.

\mr{
    Finally, the computer vision community has developed multiple approaches to anonymization that do not preserve the content of the original photo for human viewers -- including some that apply adversarial modifications~\citep{li2019anonymousnet}.
    Those approaches are best used when stronger privacy guarantees are required, such as when humans -- and not just facial search services -- are not supposed to be able to deidentify the individual in the photo. 
    Therefore, we believe they are orthogonal to this work, as we aim to allow individuals to continue to derive utility from their facial photos. 
}

\section{Goals and Assumptions}
\label{sec:goals_and_assumptions}
Our objective is to \mr{prevent previously scraped public photos of social media users from being useful to facial search services by poisoning the database of facial images}. 

In striving for this goal, we make the following assumptions, which we believe correspond to the real-world deployment scenario of such face recognition services.

\heading{The Face Recognition Service is the Adversary}
We treat the company supplying face recognition technology --- one that scrapes public photos of users from various social media websites --- as the adversary.
The source of such scraped photos may include such services as Facebook, Twitter, LinkedIn, Venmo, and others. 
This adversary records which account each photo came from or who was tagged so that the lookup photos can be linked to identities or the account that the photo was tagged for, if available.
When a third party performs a face lookup through the adversary, the system processes the query photo with a neural network, computes the photo's closest neighbors, and returns the accounts associated with those. 
In this model, we do not restrict how the third party obtains the query photo; it could be an untagged photo from a different social media company (one not scraped by the adversary), from a surveillance camera photo, or from other sources.
\mr{
    Importantly, we assume this scraping has already taken place for millions of users and is ongoing for others and for future photos of those already in the database.
    Our solution seeks to improve privacy, given that the adversary possesses some fixed amount of unmodified photos associated with individuals and that the adversary will only pick up modified photos of individuals participating in our scheme once our solution is fully deployed.
}

\heading{Social Media Users Seek Privacy}
Users of the platforms enumerated above seek to frustrate the search by ensuring that links to their profiles are not returned when the query photo truly is of them.
Where that is not possible, users prefer that many other identities are returned by the search so that theirs does not stand out.
Users may collaborate to achieve this goal and the platforms hosting the photos might also participate in the privacy enhancing scheme. 
We discuss different collaboration models in section~\ref{sec:solution_design}.

\heading{No User Control over the Query Photo}
Crucially, we assume that individuals do not control the query photos that malicious parties might submit to the face recognition service to identify them.
Individuals may not be in full control of their appearance whenever photos of them might be taken in public.
In addition, they might not wish to permanently modify their physical appearance whenever they are in public spaces, but might be willing to participate in a scheme such as ours that involves digital modifications that do not lower the quality of their digital photos. 

\heading{Limited Control over the Lookup Set}
Individuals have the ability to control \mr{future} photos that get scraped by the facial lookup service because they control the photos they upload to social media.
It is useful to distinguish between two types of individuals here.
One is individuals who do not have photos already scraped by the adversary.
This might be because all of their photos were private or because they never uploaded any photos in the first place.
Another is individuals who already have images in the adversary's database.
These individuals can begin participating in our privacy protection scheme and modify their future uploads, which the adversary then scrapes.
However, they cannot modify the photos that were previously scraped.
Thus, the adversary possesses a ``core'' set of clean images for those individuals.
\mr{
    Untagging, delisting and otherwise hiding previously scraped photos is unlikely to be an effective protection for these people, as the links between their images and their profiles already exist in the adversary's database. 
    Our solution aims to increase the privacy of this second group of individuals.
}

\heading{Access to the Model}
We assume that the protectors have access to the adversary's model and weights so that they can perform so-called ``white-box'' adversarial examples modifications. 
This assumption is not unreasonable by itself, as models often leak even from highly secure organizations.
\mr{
  In this particular scenario, the adversary may even be forced by regulators to release the model publicly for accountability and transparency purposes.
}
It is also possible that the adversary is outright using a public face recognition model that the protectors also have access to.
Without such a level of access, protectors can rely on the transferability property of adversarial examples to carry out their attacks.
\mr{
    We study how protectors can adapt their decoy generation and the effects on our scheme's privacy protections in Section~\ref{sec:transferability}.
}

\heading{No Quality Degradation of Social Media Photos}
We wish to apply a privacy defense mechanism that does not degrade the quality of photos that users post.
Legitimate human users should still be able to recognize people they know in modified photos.
Any modifications introduced to encumber computational processing of facial images should not impede human understanding.
Our system provides a tunable knob for defense, whereby tuning the knob for increased privacy can lead to more visual artifacts.
The knob settings we consider in this paper are still effective for privacy, even though they introduce only minor artifacts. 
Although outside the scope of this paper, a user study could evaluate the visual impact of these artifacts, for large knob settings.
\mr{
  For one such existing study, we refer readers to~\citet{gao2020face}.
}

\section{The FoggySight Design}
\label{sec:solution_design}
As facial lookup is primarily enabled through DL algorithms, we propose using adversarial examples\footnote{
Recall that for FoggySight, the adversary is the facial lookup service.
The ``adversarial'' designation in ``adversarial examples'' refers to adversaries against the neural network model.
In our case, the adversaries against the neural network model are the users seeking privacy from their adversary --- the facial lookup service.
}
for providing privacy for social media users. 
These are modifications to photos that shift the output of neural networks according to the modifier's choosing. 
Usually, such ``adversarial'' changes are imperceptible to humans, making them particularly attractive tools for our use case.

While the generation of adversarial examples has been well-studied in the literature, we explore how they can be used for privacy enhancements.
Thus, we explore questions around picking adversarial targets and coordination among users in doing so to achieve their privacy defense goals. 
\mr{
    Instead of focusing on how the outputs of a model or a face recognition service are affected by individual adversarial examples, we consider the broader facial search process and optimize for privacy in the recall set.
    That is, images associated with the true protected individual in the lookup set should not be returned when the service is queried with their photo or they should be returned only along a multitude of other identities.
}

In order to discuss how we generate adversarial examples, we will first introduce some notation. For specific identities $i$ and $j$, we will denote the photos that depict identity $i$ and $j$ in the lookup set as $L_i$ and $L_j$ respectively. We will use $x_i$ and $x_j$ to denote elements of $L_i$ and $L_j$, and $q_i$ and $q_j$ to denote query photos depicting identities $i$ and $j$, respectively. With this terminology, we can summarize the face recognition pipeline as follows:
\begin{enumerate}
    \item The face recognition company scrapes a lookup set from publicly available sources and obtains a trained network $f$. 
    \item The user of the face recognition technology takes a query photo $q_i$ of some identity $i$.
    \item The face recognition technology computes the top $k$ recall set $N(q_i, k)$ with respect to $q_i$ and returns them to the user (i.e., the adversary in our model), along with associated links or profiles associated with those photos in $N(q_i, k)$.
    \item The user (the adversary in our model) manually examines the set of identities in $N(q_i, k)$ and uses their own judgment to recover the true identity of the person depicted in $q_i$. If many of the photos in $N(q_i, k)$ are also in $L_i$, then the user will be able to match $q_i$ to the identity $i$.
\end{enumerate}

With this in mind, the goal of our adversarial examples is to prevent many of the photos in the set $L_i \subset L$ from being in $N(q_i, k)$.

\subsection{Overview}
\label{sec:herd_defense}
As we discuss in Section~\ref{sec:goals_and_assumptions}, an individual $i$ that cannot modify all of their photos in $L_i$, for example because they have already been scraped by the adversary. 
Clean photos in $L_i$ will be close to future query photos $q_i$ and will likely be contained in $N(q_i, k)$, thus deanonymizing the individual.
Consider Fig.~\ref{fig:soloaction}: if a blue dot were left behind, it could be used to identify the individual depicted in the query photo.
(For an empirical evaluation of this intuition, see Appendix~\ref{sec:solo_defense}.)

In order to protect privacy in this scenario, we propose instead to ``crowd out'' as many of the clean lookup set photos as possible. 
That is, we propose embedding as many decoy photos as possible with different identities into the embedding space near an individual's clean lookup photos such that those decoy photos show up in future queries, rather than the lookup photos themselves. 
If the lookup set contains many photos of \textit{other} identities that are closer to a future query photo than the clean lookup photos are, then the search will fail to recover who is truly depicted in the query. To better describe how this scheme operates, let us introduce several new terms:
\begin{itemize}
    \item \textbf{Protected users}:
    Those are users whose identity the scheme aims to protect or hide from the facial lookup.
    \item \textbf{Protectors}:
    Those are the users who choose to volunteer photos for the crowding out effect.
    By volunteering these photos, they achieve minimal additional privacy for themselves (similar to the privacy benefits from the ``solo action'' solution).
    However, they contribute to the privacy of protected users.
    Users can be both protected and protectors but we highlight these different groups to show that the benefit is concentrated on the protected whereas the action is needed from the protectors.
    \item \textbf{Decoy photos}:
    Photos that depict the protectors in reality but for which neural networks produce embeddings in the region of the protected.
    We aim to ensure that decoy photos --- as opposed to clean photos of the protected --- are returned in the recall set in response to a query.
\end{itemize}

\begin{figure}[tb]
    \centering
    \includegraphics[width=0.9\columnwidth]{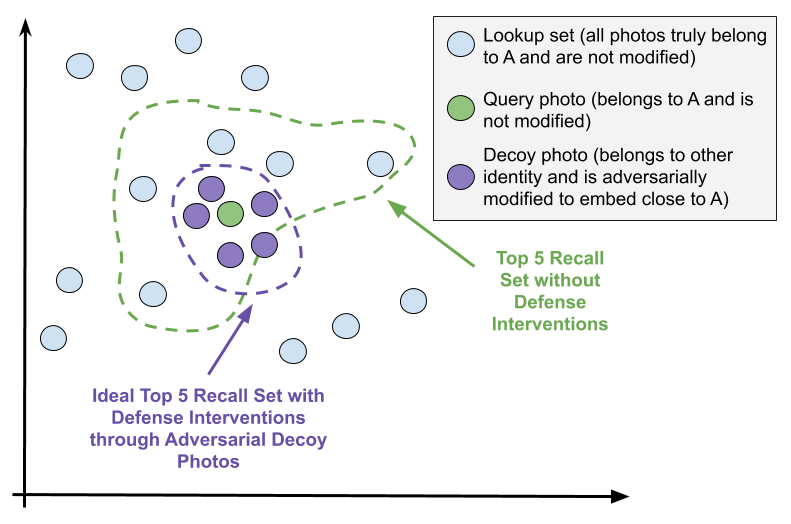}
    \caption{ 
    Visual illustration of the FoggySight privacy defense strategy. 
    Decoy photos are  pictures belonging to different identities that are adversarially modified so that face recognition neural networks produce embedding vectors close to those of the identity being protected (denoted as ``A'').
    Therefore, decoy photos appear as the closest neighbors of a query photo of A and the real identity is not revealed in response to the query. 
    }
    \label{fig:generalapproach}
\end{figure}

With these terms in mind, the scheme operates as follows:
\begin{enumerate}
    \item Protectors create decoy photos by means of adversarial examples-generation algorithms.
    \item Protectors upload those photos to their social media profiles and make them public.
    \item The adversary (the facial lookup company) scrapes those decoy photos and pre-computes their embeddings, as usual for all photos.
    \item When a query is run on a protected identity, the closest matches are decoy photos belonging to a different identity.
\end{enumerate}
For a visual representation of this idea, see Fig.~\ref{fig:generalapproach}.

\subsection{Adversarial Examples Generation}
\label{sec:advexgen}
To generate targeted adversarial examples, given a face recognition model $f$, a target vector $v \in \mathbb{R}^d$ and an image $x \in \mathbb{R}^{w \times h \times c}$ users can solve this optimization problem for an adversarial perturbation $\delta$:
\[\arg\min_{\delta} D\left(f(x + \delta), v\right) \textrm{ such that } ||\delta||_\infty \leq \epsilon \;. \]
This can be solved with projected gradient descent, as proposed in~\cite{madry2017towards}. 
Note that $\delta$ is unique to each image $x$. 
Also, note that this optimization procedure may not converge ideally and there might be a gap in the vector space between the target $v$ and $f(x+\delta)$. 
In the next section, we discuss how pairs of $x$ and $v$ are to be selected for maximum effectiveness of the strategy.

\mr{
    In some cases, the face recognition model $f$ that the protectors have access to may not match the model that the facial search provider uses. 
    For those situations, the protectors can generate \textit{robust} adversarial examples by applying the Expectations-over-Transformations (EOT) algorithm~\citep{athalye2017synthesizing}. 
    This boils down to solving the following optimization objective:
    \[\arg\min_{\delta} \mathbb{E}_i \left[ D\left(f(T_i(x + \delta)), v\right) \right] \textrm{ such that } ||\delta||_\infty \leq \epsilon \;  \]
    where $T_i$ are image transformations -- such as cropping, brightness shifts, additive Gaussian noise, etc -- applied with randomly sampled parameters.  
    In practice, we solve this objective by drawing the parameters for the transformation randomly at each step of the projected gradient descent algorithm. 
    This boosts transferability of adversarial examples because they acquire more universal features that two different models learn to use in computing their predictions. 
    In our experiments, we use random brightness shifts, random cropping, and additive Gaussian noise for this purpose.
    We give the exact parameters of our optimization runs in Appendix~\ref{sec:experiment_parameters}.
 }
 
 \mr{
    A further way to boost transferability is to generate decoys against an ensemble of face recognition models (see~\citep{liu2016delving}). 
    This works by solving the following objective for models $f_1, ...,f_j, ..., f_n$: 
    \[\arg\min_{\delta} \mathbb{E}_i \left[ \sum_{j=1}^n D\left(f_j(T_i(x + \delta)), v\right) \right] \textrm{ such that } ||\delta||_\infty \leq \epsilon \;  \]
}

\subsection{Selecting Targets}
For an individual with identity $i$ and lookup photos $L_i$, the overall goal is to have others embed many decoy photos near the photos in $L_i$ such that a new query photo's neighbor set $N(q_i, k)$ contains mostly decoy photos rather than photos from $L_i$. Ideally then, the targets $v$ chosen for adversarial example generation should be embedding vectors corresponding to photos that either belong to $L_i$, or points close to such vectors. In this section we enumerate several different strategies for picking such targets.

\label{sec:selecting_targets}
\heading{Same Universal Target} 
First, all users contributing decoy photos could select a single photo of the protected user and modify all of their images so that they embed close to that one photo. 
This has two benefits: simplicity and an extra layer of privacy for the defended individual.
When everybody creating decoy photos has the same target, there is no problem of coordination.
Everybody knows exactly how to modify their photos and does not need to check with anybody else in the scheme on what target to use.
Such a mechanism also reveals the least amount of information about the protected user.
This is particularly important as previous work has established that facial embedding vectors can be reversed to obtain the original appearance of the individual~\cite{fredrikson2015model}. 

Unfortunately, this poisoning scheme is unlikely to be very effective. 
A single sample from the distribution of photos of the defended individual is probably not a good representative for the entire distribution.
If the defenders are ``lucky'' and this is the most probable sample, then many other lookup images will be crowded out by the decoys.
However, if they are not, the crowding out effect will be limited as the query photo is likely to land far away from all the decoys and closer to other clean images of the target.

\heading{Randomly Sampled Lookup Set Photo as Target} 
As a second  approach, each user in the decoy-generating group could pick a random lookup set photo of the protected user as their target.
The benefits of this scheme is that with large enough numbers of decoy photos, the community can easily crowd out every single lookup set photo of the user. 
In fact, if the run of the adversarial examples generation algorithm converges perfectly, then a linear number of decoy photo is sufficient to crowd out the clean photos, no matter where the query photo lands. 
This will happen because the closest photo --- along with its decoys --- will fill the search result set (assuming the embeddings of the decoy photos land exactly on top of the clean photos). 

Unfortunately, adversarial examples algorithms do not converge perfectly in practice. 
Thus, to achieve perfect crowding out, the final error of the decoy photos needs to be in a favorable direction to the defenders (the decoy photos need to land between the query photo and the lookup set photos). 
Since neither the exact position of the query photo nor the error in the adversarial examples generation are easily predictable, the scheme might need more than a linear number of decoys to achieve its goals.
Even worse, this targeting approach requires honest cooperation by all defenders in drawing the target photos uniformly at random.
Any intentional or unintentional bias in the selection of the targets (a deviation from the uniform sampling) for the decoys reduces this scheme's effectiveness.

The deficiencies of the solutions proposed above reveal that the community generating decoy photos should make use of the fact that the defenders know the structure of the lookup set a priori.
They can take advantage of this fact in two ways.
First, they could use the lookup set to estimate the most likely point where the query photo will land.
Then, the decoy photos could be concentrated in that region.
Alternatively, they could attempt to distribute the decoy photos so that they are closest to the lookup set photos that are highest likelihood. 
We present instatiations of these ideas next.

\heading{Targeting the Mean Vector of the Lookup Set}
Assuming that query photos and lookup set photos are drawn from the same distribution and that it is sufficiently similar to the normal distribution, the most likely point for the query photo to land on is the mean of that distribution.
This is easily estimated with the mean of the lookup set, assuming it is sufficiently large.
Further, if the variance of the distribution of photos of the same identity is low, the identity photo is unlikely to be far away from the mean.
Therefore, a large concentration of decoy photos around the mean should easily crowd out most lookup set photos.

\heading{Targeting a Sample from a Fitted Distribution} 
Another conjecture is that the distribution that query photos are drawn from might have higher variance than the distribution of the lookup set (but the same mean).
Certainly, this is possible as query photos are likely to be sourced from uncontrolled enviornments that might be very different from the social media photos used to build up the lookup set (e.g., CCTV). 
In this situation, it is preferable to introduce decoys that do not land exactly on the mean of the lookup set. 
We explore drawing targets from a Gaussian distribution with mean and variance matching that of the lookup set.

\subsection{Collaboration Models}
Regardless of the targeting strategy, protectors need to collaborate in order to achieve maximum effectiveness. 
We describe collaborations ranging from no collaboration at all to a fully decentralized approach with everyone participating. 

    \heading{No Collaboration}
    In this setting, protected users are their own protectors.
    They can flood Internet websites that are to be scraped or create fake accounts with decoy photos.
    A limitation of this approach is that a user acting alone is unlikely to be able to generate enough decoy content without violating other policies.
    
    \heading{Centralized Assignment}
    In this setting, a trusted central party (e.g., a social media company) endeavors to protect the privacy of its own users from facial lookups.
    The company could apply all alterations automatically to users who opt in or to all users by default. 
    This has the benefit that users need not coordinate or trust each other at all.
    
    The company can make centralized decisions for targeting and adapt the scheme as necessary.
    The problem with this model is that the solution is not platform-agnostic and users can still be deanonymized from photos on websites that do not apply this protection.

    \heading{Decentralized Collaboration}
    Users can collaborate with each other to select targets and modify their photos.
    This could be mediated by a browser extension or a phone app that automatically applies the needed  modifications for the photos to act as decoys. 
    Indeed, this approach does not require the consent of the protected individual at all, as protectors could even scrape the protected's public photos themselves.
    The downside to this approach is that coordination is difficult.
    Protectors may not follow the protocol correctly, they might be running outdated versions of the software, they could outright go rogue and pick arbitrary targets or not participate at all. 
    Protectors also might not be aware if they are picking decoy photos of other protectors or if they are using the clean photos as targets.
    We explore issues arising from this difficulty in coordination in Appendix~\ref{sec:community_naive_random_iterated}.

\mr{
    \subsection{Matching Protectors and Protected}
    In all cases, the matching of protected and protectors need only follow a simple rule:
    No single protector should provide too many of the decoys for a given protected individual, relative to the number of decoys by other protectors.
    To illustrate why this rule is important, consider an extreme scenario with only one protector for a given protected individual.
    When a query is run for the protected individual, the single protector will appear as the most likely individual whose face belongs to that user.
    This means that the protector will now suffer whatever negative consequence were targeted at the protected.
    By contrast, if many different protectors are returned, then the facial search user will not be able to identify any individual in the query photo (mistakenly or otherwise) with any reasonable degree of certainty.
    This is captured by our ``identity uniformity'' metric (see Sections~\ref{sec:experimental_setup_and_metrics} and~\ref{sec:evaluation}).
    Beyond this rule, FoggySight is agnostic to how protectors are matched up with protected users.
}

\section{Experimental Setup and Metrics}
\label{sec:experimental_setup_and_metrics}
In the experiments that follow, we aim to study and understand which strategy performs best in terms of protecting individual privacy. 
In order to do so, we need to define quantitative metrics that represent success when it comes to privacy protection.

\mr{\subsection{Metrics}}
The first metric we call \textit{recall percentage at k}.
Intuitively, it is defined as the percent of the target's photos that appear in the top $k$ matches from the lookup set. 
This is meant to reflect a scenario in which the user of a face recognition system has a limited ability to look through the top $k$ matches. 
It is formally defined as:

\begin{equation}
\textrm{RP}_k (A, q_i) =
 \frac{\sum_{x \in N(q_i, k) } \mathbb{I}[x \in L_i]}{k}
\end{equation}

\noindent
where $q_i$ is a query image depicting individual $i$, $L_i$ is the photos in the lookup set that also depict individual $i$, and $\mathbb{I}$ denotes the indicator function. We assume that procedure $A$ has been used to modify some portion of the images in the lookup set $L$. 

The second metric we call \textit{discovery rate at k}.
Intuitively, it is defined as the percentage of the time that any photo from the target identity appears in the top $k$ matches from the lookup set. This is meant to reflect the scenario in which the user of the face recognition system has the resources to look through and investigate every single photo in the top k matches. Formally, we define it as:

\begin{equation}
    \textrm{DR}_k(A, q_i) = \mathbb{I}
    \left[
        \exists x \in N(q_i, k) \textrm{ such that } x \in L_i \right] 
\end{equation}

\noindent
That is, it is 1 if there exists at least one photo of individual $i$ in the neighborhood around $q_i$. Although the discovery rate for a single image $x$ is either 0 or 1, we can take the expectation over many images from a single identity to get the expected discovery rate for that identity, or over all images in $L$ to get the expected discovery rate for the adversarial procedure $A$.

The third metric we call \textit{identity uniformity at k}. 
Intuitively, it captures how many different identities are present in the recall set (subject to normalization).
Lower identity uniformity (close to 0.0) means that every possible identity is included in response to a query.
Thus, privacy is protected because the privacy adversary cannot be reasonably certain which identity of all the possible ones is depicted in the query (it could be any of them).
Higher identity uniformity means the privacy adversary can reasonably examine all returned identities closer to violate the privacy of the person in the query. 
Formally, we define identity uniformity for a query photo $q_i$ as:
\begin{equation}
   \textrm{IDUnif}_k(q_i) = 1 - \frac{ID(N(q_i, k))}{ID(L)}
\end{equation}
where $ID$ is a function that maps a set of images to the number of unique identities depicted in those images. As with recall and discovery, we take the expectation over all photos serving as queries.

\mr{\subsection{Dataset and Models}}
In our exploration, we use the VGGFace2 test dataset~\cite{cao2018vggface2} \mr{for evaluation}.
In order to make the exploration tractable given limited computational resources, we sampled 19 identities and 50 photos of each uniformly at random and performed all experiments on them.
\mr{The original test dataset that we sample from has 500 identities. The full VGGFace2 dataset consists of 9,000 identities total with an average of 362 faces per subject.}
During our explorations, we additionally ran \mr{some of our} experiments on the full dataset and did not find the results to be substantially different.
Therefore, we believe the results that we present are more broadly applicable, despite the subsampling.
\mr{
    To perform our experiments, we modify each of the 50 photos of each of the 19 identities 18 times -- one for each other identity -- by using the algorithms and targeting schemes set out in Section~\ref{sec:advexgen}. 
    The 50 clean photos of each identity are also used to compute the targets as set out in Section~\ref{sec:selecting_targets}.
    Then, we sample from the resulting decoys and from the original (subsampled) set of clean photos to build up a lookup set. 
    This set corresponds to the poisoned dataset that the facial search system would scrape from the Internet to provide its service.
    Query photos are selected at random from the remaining photos (that were not included in the 50) in each identity to simulate an image that was taken of the target in public.
    All metrics reported are averaged over multiple query photos.
}

\mr{
    We perform all experiments on the Inception-ResNet v1 network~\cite{szegedy2016inception} trained on the VGGFace2 training set~\cite{cao2018vggface2} and originally implemented at the following GitHub repository: \url{https://github.com/nyoki-mtl/keras-facenet}, which is itself a reimplementation of this repository: \url{https://github.com/davidsandberg/facenet}. 
    For transferability experiments (Section~\ref{sec:transferability}), we use the original implementations at \url{https://github.com/davidsandberg/facenet} and use a second network with the same architecture but trained on the Casia-Webface dataset~\citep{yi2014learning}.
    We do not process any images from the Casia-Webface dataset but merely use the pretrained network.
}

\mr{
    We also study transferability to the Microsoft Azure Face API service available here: \url{https://azure.microsoft.com/en-us/services/cognitive-services/face/}.
    This service allows its users to specify a training set of images associated with a set of identities.
    For this purpose, users create ``person groups.''
    These person groups are loaded with images for each person and then trained, but the documentation does not provide details on what kind of model is used for this purpose. 
    When a person group is queried, the service responds with the identity of the person it believes is in the photo or with an empty response if it does not identify anybody from the person group's members.
    In our measurements, we consider only a response with the correct identity as a correct response and empty responses and responses with an identity not matching the ground truth of the query photo are considered wrong.
    Thus, for experiments on the Azure Face Service, we only report the equivalent of recall at $k=1$.
}

\section{Evaluation}
\label{sec:evaluation}
\mr{
In this section, we report results of our experimental evaluation of FoggySight.
We further provide full details of our experimental parameters in Appendix~\ref{sec:experiment_parameters}.
}

\subsection{Adversarial Examples Success}
\label{sec:experiment_advex_success}
We first analyze how well the adversarial examples generation algorithm achieves its goal of shifting the output of the neural network while producing images indistinguishable from the original photo.

\begin{figure}[tb]
    \centering
    \includegraphics[width=0.75\columnwidth]{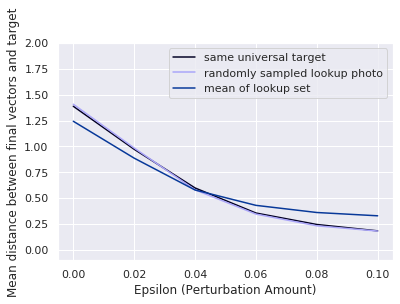}
    \caption{ 
        Magnitude of final optimization loss after decoy photo generation under different perturbation magnitudes $\epsilon$. 
        Note that the case where $\epsilon = 0.0$ corresponds to the unmodified photos.
        As expected, the higher the perturbation amount, the better the PGD algorithm for adversarial examples generation achieves its goal.
    }
    \label{fig:herd_final_loss}
\end{figure}

To do so, we begin by measuring the final distance in the embedding space between the vectors produced by the neural network for decoy photos and their respective targets.
The results are given in Fig.~\ref{fig:herd_final_loss}.

As expected, we can observe that all perturbation amounts manage to shift the output of the neural network.
Furthermore, higher perturbation amounts are more successful at bringing the final neural network loss close to their target.
Note that even at the highest perturbation amounts, there is a level of ``irreducible'' loss and the optimization algorithm does not always achieve its goal perfectly. 

\begin{figure}[tb]
    \centering
    \begin{subfigure}{.3\columnwidth}
        \centering
        \includegraphics[width=\columnwidth]{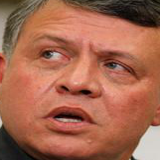}
        \caption{no modifications}
    \end{subfigure}
    \begin{subfigure}{.3\columnwidth}
        \centering
        \includegraphics[width=\columnwidth]{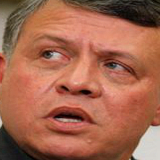}
        \caption{$\epsilon = 0.02$}
    \end{subfigure}
    \begin{subfigure}{.3\columnwidth}
        \centering
        \includegraphics[width=\columnwidth]{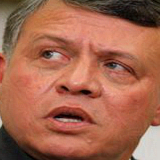}
        \caption{$\epsilon = 0.04$}
    \end{subfigure}
    
    \begin{subfigure}{.3\columnwidth}
        \centering
        \includegraphics[width=\columnwidth]{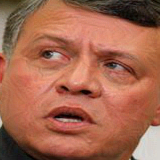}
        \caption{$\epsilon = 0.06$}
    \end{subfigure}
    \begin{subfigure}{.3\columnwidth}
        \centering
        \includegraphics[width=\columnwidth]{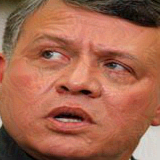}
        \caption{$\epsilon = 0.08$}
    \end{subfigure}
    \begin{subfigure}{.3\columnwidth}
        \centering
        \includegraphics[width=\columnwidth]{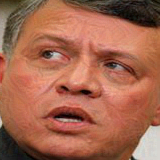}
        \caption{$\epsilon = 0.1$}
    \end{subfigure}
    
    \begin{subfigure}{.3\columnwidth}
        \centering
        \includegraphics[width=\columnwidth]{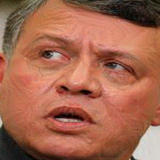}
        \caption{$\epsilon = 0.2$}
    \end{subfigure}
    \begin{subfigure}{.3\columnwidth}
        \centering
        \includegraphics[width=\columnwidth]{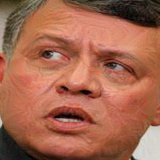}
        \caption{$\epsilon = 0.5$}
    \end{subfigure}
    \begin{subfigure}{.3\columnwidth}
        \centering
        \includegraphics[width=\columnwidth]{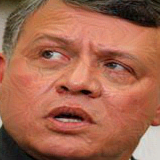}
        \caption{$\epsilon = 0.7$}
    \end{subfigure}
    
    \begin{subfigure}{.3\columnwidth}
        \centering
        \includegraphics[width=\columnwidth]{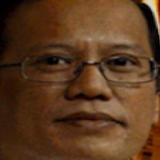}
        \caption{Target image (identity n000958)}
    \end{subfigure}

    \caption{ 
        Illustration of final decoy images under different perturbation magnitudes $\epsilon$.
        These are images of subject n000029 in the VGGFace2 dataset modified according to the ``randomly sampled target from the lookup set'' strategy to produce vectors in the region of subject n000958.
        Observe that even high perturbation magnitudes such as $\epsilon=0.2$, $0.5$, and $0.7$ --- higher than the values we find appropriate for achieving our privacy goals --- yield negligible degradation in quality.
    }
    \label{fig:herd_final_images}
\end{figure}

It is also useful to understand how these perturbations look visually. 
We show the final decoy images with different perturbation amounts in Fig.~\ref{fig:herd_final_images}.
Even high perturbation amounts do not distort the image to an unrecognizable amount. 
Therefore, we do not believe that this will have a high impact on user experience.

\subsection{Privacy Protection Success as a Function of $\epsilon$}
\label{sec:experiment_recall_vs_epsilon}

We next analyze how well the decoys fare based our metrics: recall, discovery, and identity uniformity.
In this section, we consider two parameters: the size of the recall set $k$ and the perturbation magnitude $\epsilon$. 
Note that $k$ is set by the adversary whereas the protectors get to pick $\epsilon$.
We seek to understand what $\epsilon$ achieves the optimal tradeoff between degrading the image quality and achieving the privacy protection goals under enough various settings for $k$.

\begin{figure}[tb]
    \centering
    \begin{subfigure}{0.49\columnwidth}
        \centering
        \includegraphics[width=\columnwidth]{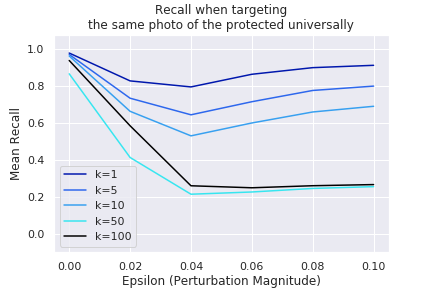}
        \caption{Recall when targeting the same photo of the protected user}
        \label{fig:recall_community_naive_same}
    \end{subfigure}
    \begin{subfigure}{0.49\columnwidth}
        \centering
        \includegraphics[width=\textwidth]{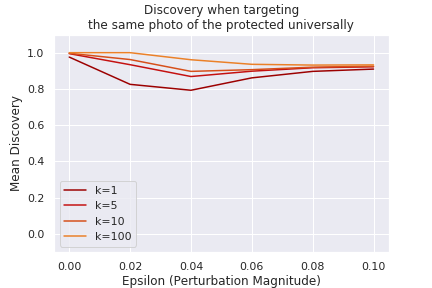}
        \caption{Discovery when targeting the same photo of the protected user}
    \end{subfigure}
    
    \begin{subfigure}{0.49\columnwidth}
        \centering
        \includegraphics[width=\columnwidth]{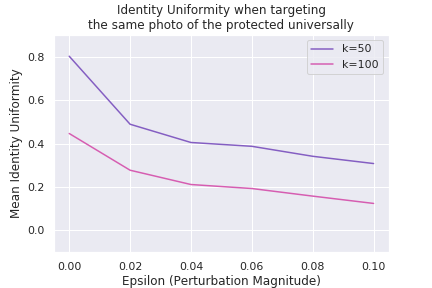}
        \caption{Identity uniformity when targeting the same photo of the protected user}
    \end{subfigure}

    \caption{
       Privacy strategy success when targeting the same photo of the protected user universally.
       All results averaged over all identities an all photos.
       While this strategy does manage to bring recall down, it is less effective at reducing the discovery rate and the uniformity of identities in the top recall set. 
    }
    \label{fig:success_epsilon_community_naive_same}
\end{figure}

\heading{Same Universal Target}
We begin with the strategy of selecting the same single photo of the protected to serve as a target for the decoys of all protectors.
The results are given in Fig.~\ref{fig:success_epsilon_community_naive_same}.
While this is the simplest strategy that exposes the least information about the protected to the protectors, these benefits come at a large cost.
We can observe that recall is only moderately impacted (an ideal protection scheme brings recall down to 0.0). 
In fact, even at high perturbation magnitudes, a photo with the real identity of the protected is the closest neighbor to the query between 80\% and 90\% of the time. (See Fig.~\ref{fig:recall_community_naive_same} and the values for recall at $k=1$. 
When the recall set contains only one photo, that photo is the closest neighbor to the query.)
The discovery rate remains consistently high for all perturbation amounts and recall set sizes, which indicates that at least one photo of the protected is available in a high pecentage of the searches ($>90\%$).

\begin{figure}[tb]
    \centering
    \begin{subfigure}{0.49\columnwidth}
        \centering
        \includegraphics[width=\columnwidth]{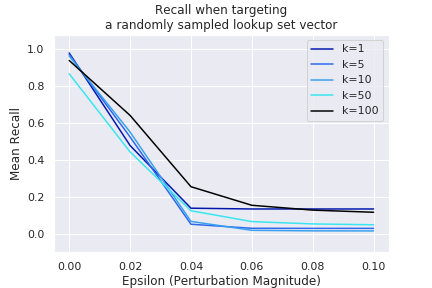}
        \caption{Recall when targeting a randomly sampled lookup set vector}
        \label{fig:recall_community_naive_random}
    \end{subfigure}
    \begin{subfigure}{0.49\columnwidth}
        \centering
        \includegraphics[width=\columnwidth]{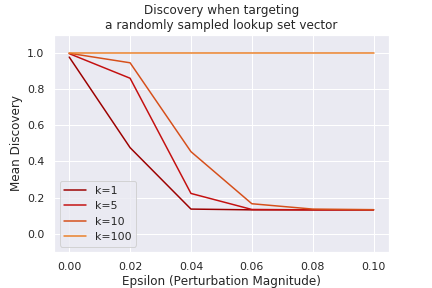}
        \caption{Discovery when targeting a randomly sampled lookup set vector}
        \label{fig:discovery_community_naive_random}
    \end{subfigure}
    
    \begin{subfigure}{0.49\columnwidth}
        \centering
        \includegraphics[width=\columnwidth]{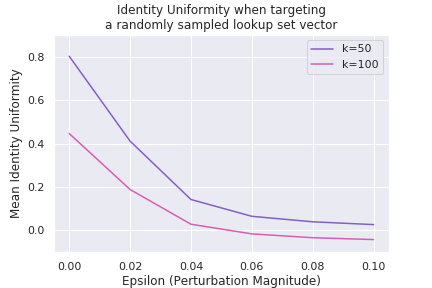}
        \caption{Identity uniformity when targeting a randomly sampled lookup set vector}
        \label{fig:numids_community_naive_random}
    \end{subfigure}

    \caption{
       Plots of privacy strategy success when targeting a randomly sampled lookup set vector.
       Observe that perturbation magnitudes of $\epsilon \geq 0.06$ achieve low recall, low discovery, and high identity uniformity, thereby successfully preserving the privacy of the protected individuals. 
    }
    \label{fig:success_epsilon_community_naive_random}
\end{figure}

\heading{Randomly Sampled Lookup Set Photo as Target}
We next turn our attention to using the entire lookup set as targets for the decoy photo optimization by the protectors.
The results are presented in Fig.~\ref{fig:success_epsilon_community_naive_random}.
It is immediately obvious that this targeting strategy performs much better. 
For $\epsilon \geq 0.04$, recall at $k=1$ is only 20\%, indicating that the closest neighbor of the query belongs to the true identity less than a fifth of the time.
For higher $k$'s, only a small percentage of the recall set ends up truly belonging to the protected identity, as can be seen by values for recall close to 0 in Fig.~\ref{fig:recall_community_naive_random}.
This success can also be confirmed by the low values for the discovery rate -- indicating that the protected identity is present in the recall set in only a fifth of the cases (see Fig.~\ref{fig:discovery_community_naive_random}). 
An exception to be observed is that the discovery rate at $k=100$ remains 100\% no matter the perturbation magnitude. 
This can be explained by the fact that at these values of $k$, the search casts a very wide net which catches at least one photo of the protected.
However, as can be seen in Fig.~\ref{fig:numids_community_naive_random}, at $\epsilon\geq 0.06$, almost all photos in such large recall sets belong to different individuals (identity uniformity is close to 0.0).
Therefore, this defense strategy successfully achieves its goal of preserving the privacy of the protected individuals.

\begin{figure}[tb]
    \centering

    \begin{subfigure}{0.49\columnwidth}
        \centering
        \includegraphics[width=\columnwidth]{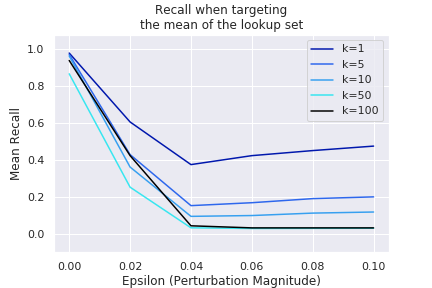}
        \caption{Recall when targeting the mean of the lookup set}
         \label{fig:recall_community_naive_mean}
    \end{subfigure}
    \begin{subfigure}{0.49\columnwidth}
        \centering
        \includegraphics[width=\columnwidth]{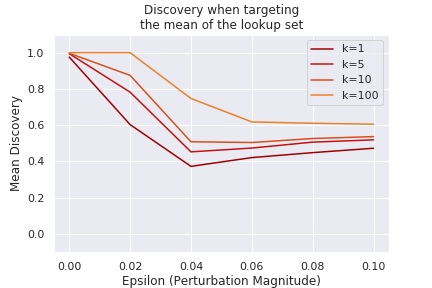}
        \caption{Discovery when targeting the mean of the lookup set.}
        \label{fig:discovery_community_naive_mean}
    \end{subfigure}
    
    \begin{subfigure}{0.49\columnwidth}
        \centering
        \includegraphics[width=\columnwidth]{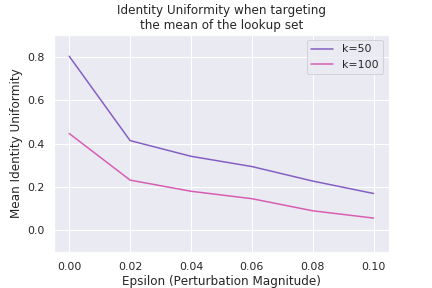}
        \caption{Identity uniformity when targeting the mean of the lookup set}
        \label{fig:numids_community_naive_mean}
    \end{subfigure}
    \caption{
       Plots of privacy strategy success when targeting the mean of the lookup set.
       While this defense leaks less information to the protectors and is easier to coordinate, it does not achieve results as good as when targeting a randomly sampled lookup set photo.
    }
    \label{fig:success_epsilon_community_naive_mean}
\end{figure}

\heading{Targeting the Mean of the Lookup Set}
While targeting a randomly sampled lookup set photo is successful, it does come with some downsides, as discussed in section~\ref{sec:selecting_targets}.
Therefore, we also experiment with using the mean of the lookup set as a target. 
Comparing every panel of Fig.~\ref{fig:success_epsilon_community_naive_mean} to every panel of Fig.~\ref{fig:success_epsilon_community_naive_random} reveals that this targeting strategy is not as effective. 
For any given combination of $\epsilon$ and $k$ values, targeting a randomly selected photo of the lookup set of the protected yields more effective decoys. 
Recall is between 10 and 20\% higher, indicating that there's more photos of the query identity being returned and less decoy photos, on average, in response to queries.
Similarly, identity uniformity rises for this same reason. 

There is one exception, however. 
For high values of $k$ (e.g., $k=100$), the discovery rate is consistently lower when targeting the mean of the lookup set (compare Figs.~\ref{fig:discovery_community_naive_mean} and~\ref{fig:discovery_community_naive_random}).
This indicates that targeting the mean does perform one function well --- it places decoy photos close to where the query photo lands in the embedding space.
Thus, as $k$ grows, less photos belonging to the protected individual are included in favor of decoy photos.  
To see this, observe that recall falls with $k$ in Fig.~\ref{fig:recall_community_naive_mean} whereas it grows with $k$ in Fig.~\ref{fig:recall_community_naive_random}.
Unfortunately, the closest photos to the query do still belong to the protected, thereby hurting the metrics for low values of $k$ (see the values for $k=1, 5, 10$). 

\begin{figure}[tb]
    \centering
    \begin{subfigure}{0.49\columnwidth}
        \centering
        \includegraphics[width=\columnwidth]{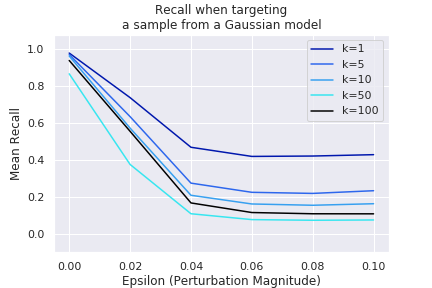}
        \caption{Recall when targeting a sample from a Gaussian model}
        \label{fig:recall_community_sample_gaussian_model}
    \end{subfigure}
    \begin{subfigure}{0.49\columnwidth}
        \centering
        \includegraphics[width=\columnwidth]{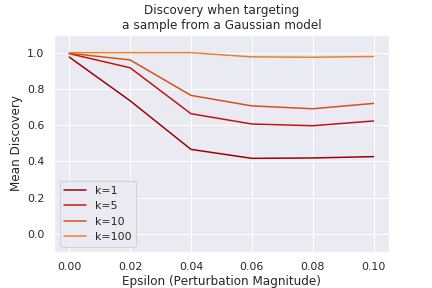}
        \caption{Discovery when targeting a sample from a Gaussian model}    
        \label{fig:discovery_community_sample_gaussian_model}
    \end{subfigure}
    
    \begin{subfigure}{0.49\columnwidth}
        \centering
        \includegraphics[width=\columnwidth]{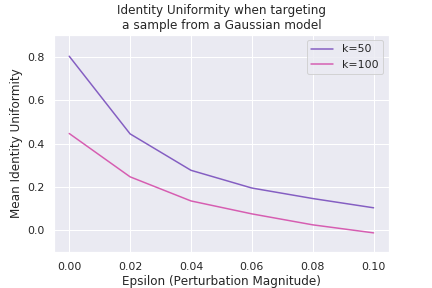}
        \caption{Identity uniformity when targeting a sample from a Gaussian model}
        \label{fig:numids_community_sample_gaussian_model}
    \end{subfigure}

    \caption{
       Graphs of privacy strategy success when targeting a sample from a Gaussian model.
       Observe that this scheme fares just as well as when targeting the mean lookup set by comparing with Figure~\ref{fig:success_epsilon_community_naive_mean}.
    }
    \label{fig:success_epsilon_community_sample_gaussian_model}
\end{figure}

\heading{Targeting a Sample from a Gaussian Model}
As another alternative, we evaluate targeting a \textit{sample} from a Gaussian model with mean and standard deviation matching that of the lookup set.
Results are given in Fig.~\ref{fig:success_epsilon_community_sample_gaussian_model}. 
The results at all settings of $k$ and $\epsilon$ are as good or worse than the results when targeting the mean.
For example, for $\epsilon=0.06$ and $k=5$, discovery (in Fig.~\ref{fig:discovery_community_sample_gaussian_model}) remains up to 10\% higher.
This is likely because the residual loss from not achieving the optimization objective perfectly introduces enough variation when targeting the mean to scatter the decoys well.
By contrast, when we purposefully introduce additional error through targeting a sample from a non-0 variance Gaussian, the decoys land farther away from the query photo.

\heading{Takeaways From All 4 Experiments}
There are several patterns to observe that are common across the 4 experiments presented in this section.
First, the higher the perturbation magnitude, the more effective the protection scheme is across all metrics and across all targeting approaches. 
More importantly, the ``optimal'' value of $\epsilon$ appears to be $0.06$ (see, e.g., Fig.~\ref{fig:discovery_community_naive_random}; the lowest discovery is achieved at $\epsilon=0.06$). Increasing the perturbation magnitude to $0.08$, or $0.1$ only improves the protection scheme by marginal amounts. 
Thus, to achieve the best tradeoff between degrading image quality and achieving the privacy goals, we recommend using $\epsilon=0.06$.

Second, at high $k$'s, it is impossible to drive the discovery rate to 0 no matter the perturbation magnitude and the targeting strategy.
This is probably because the search casts a very wide net at such high values of $k$.
However, in terms of privacy, this is not a problem.
In fact, at high $k$'s, our protection schemes manage to insert a large number of different identities into the top recall set (compare the b and c panels in the figures in this section).
When there are many different identities returned in response to a query, the person performing the search through the adversary's services does not know with any reasonable degree of confidence who is depicted in the query. 
Therefore, the discovery rate is perhaps a bit too harsh and the ultimate goal --- of preventing the identification of the person in the query photo --- is achieved.

\subsection{Privacy Protection Success as a Function of the Number of Decoy Photos}
\label{sec:experiment_recall_vs_num_decoys}

\begin{figure}[tb]
    \centering
    \begin{subfigure}{0.49\columnwidth}
        \centering
        \includegraphics[width=\columnwidth]{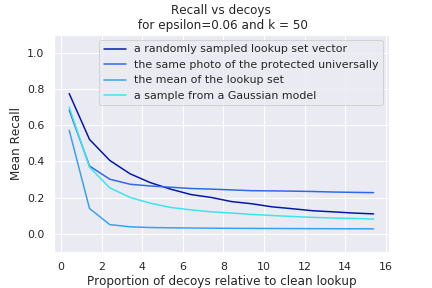}
        \caption{Recall vs. the number of decoy photos as a proportion of $k$}
    \end{subfigure}
    \begin{subfigure}{0.49\columnwidth}
        \centering
        \includegraphics[width=\columnwidth]{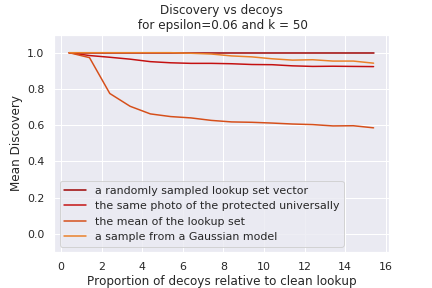}
        \caption{Discovery vs. the number of decoy photos as a proportion of $k$}
    \end{subfigure}
    
    \begin{subfigure}{0.49\columnwidth}
        \centering
        \includegraphics[width=\columnwidth]{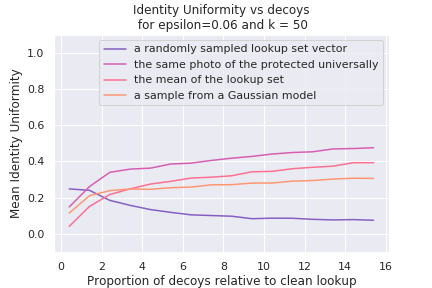}
        \caption{Identity uniformity vs. the number of decoy photos as a proportion of $k$}
    \end{subfigure}
    \caption{
       Graphs of privacy strategy success versus the number of decoy set photos.
    }
    \label{fig:success_vs_decoy_set_size}
\end{figure}

We also explored another approach to analyze the effectiveness of the different targeting strategies.
The more decoy photos are needed, the harder it is for the privacy protection to succeed.
Therefore, we ideally want a targeting strategy that achieves its goal more easily if there are less decoy photos needed.
In Fig.~\ref{fig:success_vs_decoy_set_size}, we present results for $\epsilon=0.06$ and $k=50$ on this metric.
Observe that the recall drops most quickly when targeting the mean of the lookup set.
Hence, it might be more desirable to apply this targeting mechanism with a higher $\epsilon$.
That way, the protection scheme can reap the benefits for discovery rate and identity uniformity discussed in the previous section and achieve them with less decoy photos. 

\mr{
    \subsection{Privacy Protection Success When Protectors Do Not Have Access to the Face Recognition Model}
    \label{sec:transferability}
    We also explore privacy protections with FoggySight in the scenario where the protectors do not have access to the exact face recognition model used to perform the facial search.
    As discussed in Section~\ref{sec:advexgen}, we adopt two techniques for ensuring that decoys transfer from the model they were generated with to an unknown other model: \emph{Expectation over Transformations} (EOT) for generating robust adversarial examples and ensemble adversarial examples generation. 
    In all experiments in this section, we employ the most successful method from the previous sections -- targeting the mean of the Lookup Set.
}

\begin{figure}[tb]
    \centering
        \centering
        \includegraphics[width=\columnwidth]{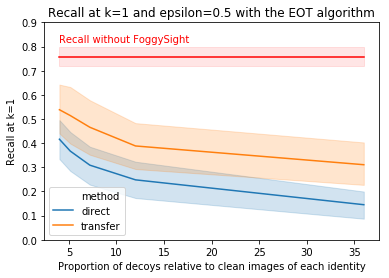}
        \caption{\mr{
        Recall at $k=1$ vs. the number of decoy photos as a proportion of the unaltered photos of an individual. 
        We have averaged and given the 95\% confidence intervals of the recall value over which network serves for generation and which one is transferred to, over all 19 identities in our subset, and over 5 different query photos not present in the lookup set (modified or not). 
        We present two results: recall when the protectors have access to the model being used (``direct'') and when they do not (``transfer''). 
        Results are plotted as a function of the number of decoys provided by protectors relative to the number of ``clean'' unmodified photos of each protected identity in the lookup set.
        We also present the recall of the network without FoggySight as a red line that does not depend on this ratio. 
        }}
    \label{fig:transfer_recall}
\end{figure}

\begin{figure}[tb]
    \centering
    \begin{subfigure}{.3\columnwidth}
        \centering
        \includegraphics[width=\columnwidth]{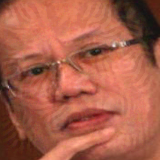}
        \caption{$\epsilon = 0.1$}
    \end{subfigure}
    \begin{subfigure}{.3\columnwidth}
        \centering
        \includegraphics[width=\columnwidth]{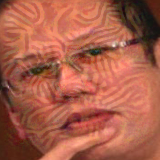}
        \caption{$\epsilon = 0.25$}
    \end{subfigure}
    \begin{subfigure}{.3\columnwidth}
        \centering
        \includegraphics[width=\columnwidth]{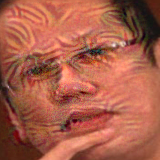}
        \caption{$\epsilon = 0.5$}
    \end{subfigure}

    \caption{ 
        \mr{
            Illustration of final transferable decoy images under different perturbation magnitudes $\epsilon$.
            These are images of subject n000957 in the VGGFace2 dataset modified to serve as decoys for other identities.
        }
    }
    \label{fig:transferable_images}
\end{figure}

\mr{   
    We first present results on transferability of decoys generated with the EOT algorithm in Figure~\ref{fig:transfer_recall} and we give sample decoy images in Figure~\ref{fig:transferable_images}.
    First observe that in both cases, the recall of the network is severely impeded both in the ``direct'' and the ``transfer'' cases. 
    The average recall drops below 0.4 with a sufficient number of decoys for both methods. 
    In other words, a protected person has less than a 40\% chance of being the nearest neighbor to their query photo -- as opposed to 90\% chance without the FoggySight defense.
    This indicates that adversarial example transferability is an effective method to poisoning the facial lookup database to increase individual privacy.
}

\mr{
    However, we also note that this defense is not 100\% effective and that there remains a gap between how effective the ``direct'' and the ``transfer'' defenses are.
    This suggests that stronger methods for generating transferable decoys are needed in order to ensure their effectiveness on unseen models.
    That is why we explore ensemble generation of adversarial examples and test the results on a commercial face recognition service -- one whose internals we do not have access to.
    In particular, we include both networks implemented in the FaceNet library for our ensemble and measure the results of the scheme on the Azure Face Recognition service.
    }

\begin{figure}[tb]
    \centering
        \centering
        \includegraphics[width=\columnwidth]{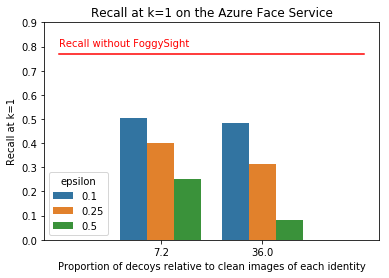}
        \caption{\mr{Recall at $k=1$ on the Azure Face Service plotted against the number of decoy photos as a proportion of the unaltered photos of the individual}}
    \label{fig:azure_recall}
\end{figure}

\mr{
    Results for this transferability to an unseen system are given in Figure~\ref{fig:azure_recall}.
    They indicate a successful scheme: when $\epsilon=0.5$ and there are 36 times more decoys than clean photos, the probability of the service identifying the protected individual is less than 10\%. 
    Therefore, FoggySight can be successful in increasing individual privacy against facial searches, even against unseen systems. 
}

\mr{
    \section{Discussion}
    \label{sec:discussion}
}
\mr{
\textbf{Practical Deployment Considerations} 
    The major step necessary for the effectiveness of FoggySight is wide community adoption.
    Our experiments -- though with a limited set of identities -- show that FoggySight requires at least  5 times more decoys than the number of unaltered photos already scraped by the facial search service to reduce the occurrence of the protected identity as a nearest neighbor to the query photo (recall at $k=1$) to less than 50\%. 
    To drive that number even further down to less than 10\% on a commercial face service, large perturbation amounts and 36 times more decoys than clean photos are required.
}

\mr{
    Based on these results, we believe FoggySight is best suited when used to frustrate facial search and create plausible deniability about who a person in a query photo is.
    With enough decoys, many different identities are returned as a response to a facial search and the true one comprises a small portion of them.
    Thus, users of the facial search service cannot be sure with a high degree of confidence that the person in the query photo is any one person from the recall set. 
    This level of protection is reasonable for the individuals similar to those represented in our dataset who may wish to increase their general level of privacy.
    However, it is absolutely not sufficient for users wishing to prevent discovery.
    The best solution for those users remains to not have their photos included in the database in the first place or to prevent query photos of themselves from being useful (e.g., through blurring or more advanced obfuscation approaches). 
}

\mr{
\textbf{Facial Search Service Countermeasures (Adaptive Privacy Adversaries)} 
    Our method relies on the ability of adversarial examples to affect the output of the facial search provider's neural network model and on their ability to remain undetected.
    There has been research on providing a variety of defenses to adversarial examples.
    Some of it has shown qualified empirical success~\cite{madry2017towards}, some has provided certification guarantees about very specific adversaries~\cite{wong2017provable} and some has even focused on top-$k$ classification~\cite{jia2019certified}. 
    However, the adversarial examples research literature has also found that robust performance (on adversarial examples) often comes at the cost of clean performance (on regular test set examples)~\citep{ilyas2019adversarial}. 
    Therefore, we believe it is unlikely that robust neural networks are going to be applied at scale for facial search, as that will trade off the system's overall reliability on unprotected and protected individuals alike.
    Furthermore, it is possible that the facial search provider detects and filters some of our decoys. 
    We believe this is out of scope for our proposal and future work should aim to quantify the effectiveness of such out-of-distribution detection.
    Our scheme remains effective as long as the ratios of decoys to clean images of a given protected individual can be maintained.
}

\mr{
\textbf{Incentives and Risks for Protectors}
    In volunteering to provide decoys, protectors increase their own privacy but also take on an added level of risk.
    First, even with FoggySight, protected individuals have an incentive to modify any future public photos of themselves so that face recognition models produce embeddings away from their ``true'' region in the space.
    This helps maintain the ratio of adversarial to clean images in the facial search provider's database.
    The fewer ``clean'' images the facial search provider has, the harder it is to identify an individual.
    Thus, protected individuals continue to have an incentive to also serve as protectors for others and participate in FoggySight actively -- as opposed to merely receiving protections.
    We emphasize that this is different from the finding that individuals cannot achieve meaningful protections on their own. 
    In Appendix~\ref{sec:solo_defense}, we explored cases where individuals wishing privacy modify their future photos in an arbitrary direction and found that that is not enough to increase privacy, given a clean query photo and some clean lookup set photos.
    FoggySight suggests that they should instead modify their photos in a specific direction.
}

\mr{
    This, however, introduces a risk for the protector. 
    If a protector participates with an unbalanced number of decoys targeted at a given protected individual, the user of the facial search tool may misidentify the protected as the unbalanced protector. 
    However, this risk can be mitigated by centralized coordination among protectors so that no single one of them is providing a larger-than-average proportion of the decoys for a given protected individual.
}

\mr{
    \textbf{Untagging and Other Defenses against Facial Search}
    FoggySight is not meant to be a standalone solution.
    In fact, the less clean photos any given user has in a database, the better decoy-based protection will work for them.
    Thus, individuals wishing to increase their facial privacy should continue to untag, take down, or otherwise delist their photos from the public Internet. 
    However, we also note that none of these solutions can succeed on its own, either.
    Reports on facial search providers~\citep{hill_valentino-devries_dance_krolik_2020} suggest that millions of individuals already have faces in those databases with links to their (possibly cached) online presence.
    No amount of untagging, delisting, or removal of photos can remedy this. 
    FoggySight aims to remedy that through poisoning the database of the facial search provider and is aided by future untagging but neither solution can work on its own.
}

\mr{
    \textbf{Dataset Limitations}
    While we believe the work in this paper establishes a proof of concept for a collaborative defense approach, all our findings are subject to the limitations of our dataset.
    For reasons of constrained computational resources, we have worked with a random sample of a bigger dataset of faces that is standard in facial recognition research (see Section~\ref{sec:experimental_setup_and_metrics}) and our results inherit all limitations of the original dataset.
    Furthermore, we acknowledge that for full deployment of FoggySight, the scheme would need to undergo rigorous at-scale testing and evaluation.
    In particular, such testing needs to ensure that different populations of users are represented properly and that protections apply to every group equally well -- and especially to groups that may suffer worse consequence of diminished facial privacy than others.
}

\mr{
    \textbf{Impact of Transferable Adversarial Examples (Decoys)} 
    In our experiments, we found that FoggySight protectors need to introduce both higher-magnitude perturbations to their images and provide more decoys when they do not have access to the adversary's model.
    For example, where protectors acting with access to the facial search model needed to inject 2-4 times more decoys with $\epsilon=0.04$ than unaltered, previously scraped images of the protected, protectors need to inject 36 times more decoys with $\epsilon=0.5$ to be really effective against commercial face recognition services with unknown internals.
    This suggests that a potential policy response that may enable individuals to apply FoggySight more effectively might be to mandate disclosure of the facial search model. 
    The \emph{best} policy responses to facial search adversaries are beyond the scope of this work, but we highlight this finding as a possible remediation mechanism that may provide individuals with more agency in protecting their privacy.
}

\section{Conclusion}
\label{sec:conclusion}
Companies today are scraping photos from social media sites and are using those photos to build powerful systems capable of identifying people from newly taken photos~\cite{heilweil2020vox}. 
We, therefore, proposed FoggySight, a community-based approach for modifying future photos provided publicly on the Internet so that they crowd out previously scraped photos. 
\mr{
    Our experiments demonstrate that FoggySight can meaningfully increase privacy.
    As with any early proposal, many practical questions need to be answered for full deployment and desired effectiveness.
    However, we are convinced that this work both highlights the limitations of facial privacy protection schemes and proposes a solid basis for future work in this space to build on.
}

\section*{Acknowledgements}
The authors would like to thank members of the University of Washington Security and Privacy Lab for their thoughtful feedback on this work.

This work was supported in part by the University of Washington Tech Policy Lab, which receives support from: the William and Flora Hewlett Foundation, the John D. and Catherine T. MacArthur Foundation, Microsoft, the Pierre and Pamela Omidyar Fund at the Silicon Valley Community Foundation; it was also supported by the US National Science Foundation (Award 156525).

\bibliographystyle{plainnat}
\bibliography{example_paper}

\appendix

    \section{Privacy Protection Success When Targeting Decoys in a Decentralized Scenario}
    \label{sec:community_naive_random_iterated}
    We consider a world where protectors fail to select clean lookup set photos as the target.
    This may occur in the decentralized collaboration setting (section~\ref{sec:solution_design}) where protectors are also protected and do not reveal information to the public regarding which of their photos serve as decoys for others.
    It is easy to imagine that a breakdown in the protection mechanism may occur in such a scenario.
    Since protectors sample decoys, every one of them is unwittingly not following the protocol and some protected individuals may end up overprotected whereas others are under-protected.
    This can be thought of as a multiround case of applying the privacy protection strategy we describe.
    Before round~1, no social media users have uploaded decoy photos.
    In round~1, protectors follow the strategy from the previous sections and deploy decoy photos.
    In round~2, protectors are now protected and other, new protectors deploy decoys for them.
    The question we seek to answer here is whether this breaks the success of the scheme.
    
\begin{figure}[tb]
        \centering
        \begin{subfigure}{0.49\columnwidth}
            \centering
            \includegraphics[width=\columnwidth]{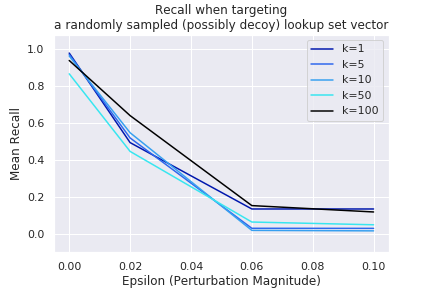}
            \caption{Recall vs. the number of decoy photos as a proportion of $k$}
        \end{subfigure}
        \begin{subfigure}{0.49\columnwidth}
            \centering
            \includegraphics[width=\columnwidth]{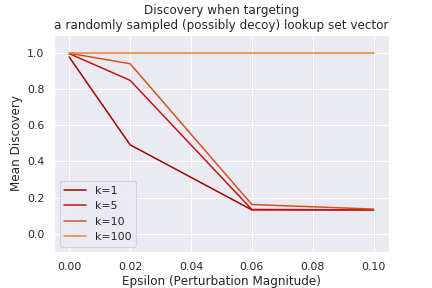}
            \caption{Discovery vs. the number of decoy photos as a proportion of $k$}
        \end{subfigure}
        \begin{subfigure}{0.49\columnwidth}
            \centering
            \includegraphics[width=\columnwidth]{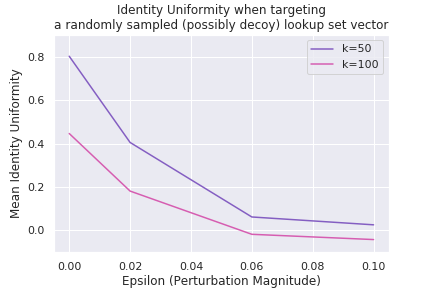}
            \caption{Identity uniformity vs. the number of decoy photos as a proportion of $k$}
        \end{subfigure}
        \caption{
           Plots of privacy strategy success when protectors target decoy photos.
           Compare to Fig.~\ref{fig:success_epsilon_community_naive_random} and note that re-targeting decoys does not negatively impact the performance metrics.
        }
        \label{fig:success_target_decoys}
    \end{figure}

    Fig.~\ref{fig:success_target_decoys} presents an experiment for round 2 where \textit{all} vectors used as targets are themselves decoys from round 1. 
    The targeting strategy used is to target a randomly sampled decoy photo (equivalent to randomly sampling a clean lookup set photo). 
    Comparing these results to Fig.~\ref{fig:success_epsilon_community_naive_random}, we find that there is no degradation in the performance metrics. 
    This effect is likely due to the fact that in round 2 all users target decoys.
    Thus, the protocol ``balances itself out'' and all protected users receive protection with decoys. 
    Note that this is only possible because the set of protectors and protected users overlap perfectly in both rounds.
    Therefore, we recommend that practical deployments of the scheme ensure that participants are both protectors and protected; otherwise, the risk of breakdown in the protection as described above remains.

\section{Solo Action Defenses with Untargeted Adversarial Examples}
\label{sec:solo_defense}
\subsection{Setup and Motivation}
Here, we consider the most natural strategy for an individual with identity $i$ trying to protect their own privacy while acting alone. 
Recall that the face recognition pipeline has a dataset of lookup photos $L$. 
Those photos in $L_i \subset L$ depict identity $i$, and correspond to photos of individual $i$ from social media websites. 
To protect their privacy, individual $i$ aims to modify the photos $x_i \in L_i$ such that $D(x_i, q_i)$ is large for some future query $q_i$. 
The individual must modify their photos prior to those photos being scraped by the face recognition system; this is a key issue that we discuss in more detail later.
Unfortunately, the individual cannot predict future query photos $q_i$. 
However, future query photos will by definition be close to the unmodified $x_i \in L_i$.
Thus, it is natural to instead modify $x_i$ to be far away from itself. We do this by solving the following optimization problem:

\[A(x_i) = \arg\max_{z} D(x_i, z) \textrm{ such that } ||x_i - z||_\infty \leq \epsilon \]

\noindent
where $x_i$ is the image that depicts individual $i$ --- one that the individual is potentially trying to upload to social media --- $A$ is the adversarial modification that transforms $x_i$, and $\epsilon$ is a pre-defined perturbation amount. This attack aims to make sure that, to the network, $x_i$ is not recognizable as the identity of the individual depicted in it, while maintaining via the constraint that it appears like a normal photo to a human observer.

To optimize this function, we use projected gradient descent, which was introduced in the context of adversarial examples by \citet{madry2017towards}. 
Although usually adversarial examples are initialized as the target image $z_0 = x_i$ doing so results in the optimization getting stuck $D(z_0, x_i) = 0$. 
We therefore follow the strategy outlined by \citet{madry2017towards} and initialize the attack with a small amount of random noise $z_0 = x_i + \hat{\mathcal{N}}_\epsilon(0, \sigma)$ for the truncated normal distribution $\hat{\mathcal{N}}_\epsilon$ truncated at $[-\epsilon, \epsilon]$.

\begin{figure}[tb]
    \centering
    \includegraphics[width=0.9\columnwidth]{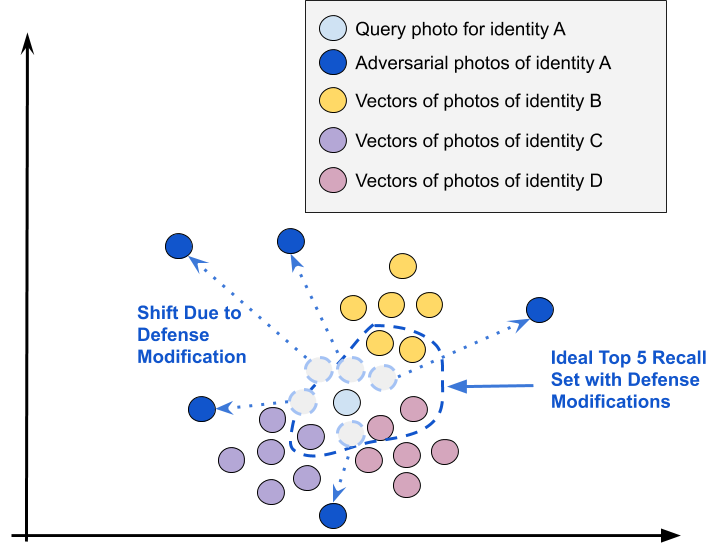}
    \caption{ A visual illustration of the solo action defense. 
    A user aims to shift his or her face images far away from their original location in the embedding space. This fills the recall set with other identities. 
    }
    \label{fig:soloaction}
\end{figure}

The motivation behind this attack is presented in Fig.~\ref{fig:soloaction}. 
By maximizing the distance in embedding space to the original, clean lookup photos, the target minimizes the chance that a new, clean photo will match any of the modified lookup set photos.

\subsection{Experimental Evaluation}
\begin{figure}[tb]
    \centering
    \begin{subfigure}{0.48\columnwidth}
        \includegraphics[width=\columnwidth]{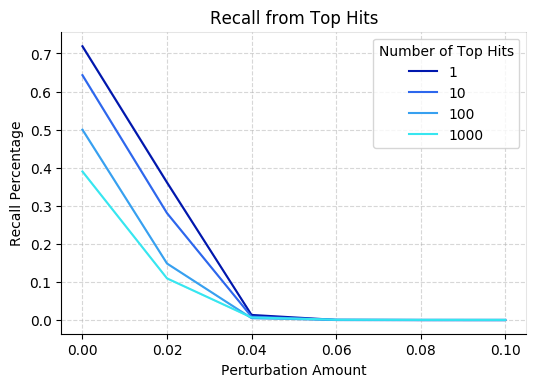}
        \label{fig:recall_self_distance_epsilon}
        \caption{Recall for the solo action attack}
    \end{subfigure}
    \begin{subfigure}{0.48\columnwidth}
        \includegraphics[width=\columnwidth]{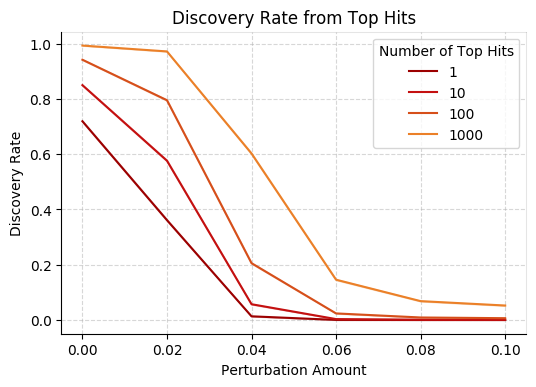}
        \label{fig:discovery_self_distance_epsilon}
        \caption{Discovery rate for the solo action attack}
    \end{subfigure}
    \caption{ 
        Recall and discovery rate at various levels of $k$ and $\epsilon$ when assuming the protected has  100\% control of their own lookup set. 
        The perturbation amount is normalized to represent percentage relative to standard deviation (images have unit standard deviation). 
        For both metrics, a perturbation amount of $0.04$ suffices to evade recognition. 
        \mr{``Top Hits'' refers to the recall set of nearest neighbors to the query photo that is returned by the facial search service to its user.}
        }
    \label{fig:sd_main}
\end{figure}

The result of applying the self distance attack to all photos belonging to the user in the database is shown in Fig.~\ref{fig:sd_main}. 
The graph averages the recall percentage and discovery rate over all images and all identities in the lookup set. 
The chart shows that perturbation amount of $0.04$ relative to an image standard deviation of 1 suffices the drop the recall percentage to almost 0 and the discovery rate at $k = 100$ to approximately 10\%. 

\begin{figure}[tb]
    \centering
    \begin{subfigure}{0.48\columnwidth}
        \includegraphics[width=\columnwidth]{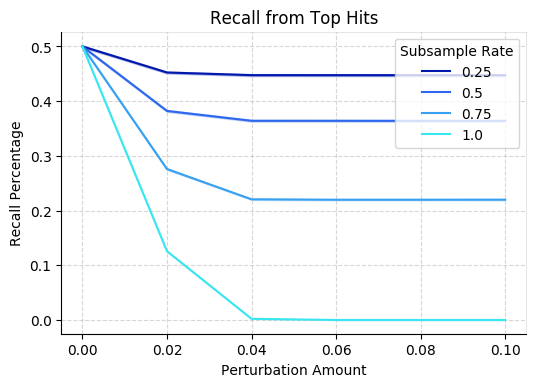}
        \caption{Recall for the solo action attack with limited control of the lookup set}
        \label{fig:recall_self_distance_subsample}
    \end{subfigure}
    \begin{subfigure}{0.48\columnwidth}
        \includegraphics[width=\columnwidth]{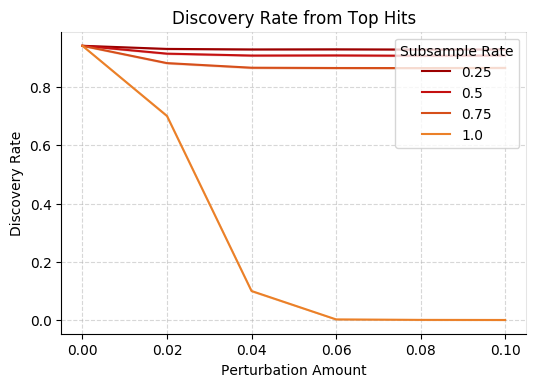}
        \caption{Discovery rate for the solo action attack with limited control of the lookup set}
            \label{fig:discovery_self_distance_subsample}
    \end{subfigure}
    
    \caption{ 
        Recall and discovery rate at various levels of $k$ and $\epsilon$ when assuming the protected only has limited control of their own lookup set (as controlled by the subsample rate). 
        The perturbation amount is normalized to represent percentage relative to standard deviation (images are have unit standard deviation). 
        Only having access to a fraction of the lookup data drastically degrades privacy protection. This indicates that other strategies are needed in the case that we cannot modify 100\% of the target's data. 
        \mr{``Top Hits'' refers to the recall set of nearest neighbors to the query photo that is returned by the facial search service to its user.}
        }
    \label{fig:sd_subsample}
\end{figure}

However, it is not always reasonable to assume that users control 100\% of their photos in the database.
Therefore, we next study the performance of the solo action attacks if only the target can only modify some fraction of their photos in the lookup set. 
We define the \textit{subsample rate} as the percentage of the target's photos in the lookup set that the target can modify. 
That is, if the adversary has 100 photos of the target in their lookup set, and the target can modify 70 of them, then the subsample rate is 70\%. 
We plot the result of subsampling using the self distance and target pair strategies in Fig.~\ref{fig:sd_subsample}. 
The plots show that subsampling even at a rate of 75\% drastically increases the expected discovery rate, which indicates that face recognition systems need only a few photos of a target out of hundreds of thousands in order to identify them. 
This indicates that our proposed attacks may not be effective enough in the case that the adversary has photos of the protected that the protected cannot modify. 
In this case, different strategies that involve many protectors acting in coordination may be needed.

\section{Further Experimental Details}
\label{sec:experiment_parameters}
For the results in Sections~\ref{sec:experiment_advex_success},~\ref{sec:experiment_recall_vs_epsilon}, and~\ref{sec:experiment_recall_vs_num_decoys}, we use a learning rate of $\alpha=0.1$ and batch size of 128, and run PGD for up to 400 iterations. 
We interrupt the optimization if the loss value has not declined for 10 consecutive iterations.
$\epsilon$ is set as indicated in the figures.
Experiments in these sections are implemented in Tensorflow 2.0~\cite{tensorflow2015-whitepaper} and use the network provided at \url{https://github.com/nyoki-mtl/keras-facenet}.

For the results in Section~\ref{sec:transferability}, we use $\alpha = 0.01$ and run the PGD algorithm for 2000 iterations without early stopping. 
We apply the following transformations with parameters sampled at random at each gradient step:
\begin{itemize}
    \item random flip left or right
    \item random brightness shift by up to $0.25$
    \item random crop of a rectangle of size $150 \times 150$, with resizing to the network input size of $160 \times 160$
    \item additive Gaussian noise with $\mu=0.0$ and $\sigma=0.5$
\end{itemize}
Experiments in this section are implemented in Tensorflow 1.15~\cite{tensorflow2015-whitepaper} and use the Inception ResNet-v4 networks implemented at \url{https://github.com/davidsandberg/facenet} and trained on VGGFace2~\cite{cao2018vggface2}  and Casia-Webface~\citep{yi2014learning}.

In order to be able to carry out experiments in a reasonable amount of time, we have sampled 19 identities uniformly at random from the VGGFace2 test dataset.
Those identities are as follows:
\begin{itemize}
    \item n000958 
    \item n001683 
    \item n001781 
    \item n002503 
    \item n002647 
    \item n002763 
    \item n003215 
    \item n003356 
    \item n004658 
    \item n005303 
    \item n005359 
    \item n005427 
    \item n007548 
    \item n008613 
    \item n008655 
    \item n009114
    \item n009232
    \item n009288
    \item n000029
\end{itemize}
We have further sampled 50 photos from each identity to include in our lookup sets and to serve as the basis for generating decoys.
This list of 1,000 photos is too large to include in the appendix, but is available upon request.
During evaluation, we sample another set of 5 photos (distinct from the 50) and use them as ``query photos.''
All metrics reported are averaged over each of these 5 photos for each of the 19 identities.

\end{document}